\documentclass{trbunofficial}
\usepackage{graphicx}
\usepackage{xcolor}
\usepackage{booktabs}
\usepackage{graphicx}
\usepackage{adjustbox}  
\usepackage{wrapfig}
\usepackage{comment}
\usepackage{silence}
\usepackage[compatibility=false]{caption}
\usepackage{subcaption}
\usepackage{placeins}   

\usepackage[hidelinks]{hyperref}
\AuthorHeaders{Saba, Khan, Ahmad, Cao, Rahman, Zhao, Huynh, and Ozguven}
\title{Infrastructure Sensor-enabled Vehicle Data Generation using Multi-Sensor Fusion for Proactive Safety Applications at Work Zone} \author{
  \textbf{Suhala Rabab Saba}\\
  Department of Civil, Construction \& Environmental Engineering, The University of Alabama\\
  Smart Communities and Innovation Building (SCIB), 28 Kirkbride Lane,\\
  Tuscaloosa, AL 35487-0288\\
  Email: ssaba@crimson.ua.edu \\
  \hfill\break
  \textbf{Sakib Khan, Ph.D.}\\
  Principal Intelligent Transportation Systems Engineer, MITRE Corporation\\
  7525 Colshire Dr, McLean, 22102 Virginia\\
  Email: sakibkhan@mitre.org \\
  \hfill\break%
\textbf{Minhaj Uddin Ahmad}\\
  Department of Civil, Construction \& Environmental Engineering, The University of Alabama\\
  Smart Communities and Innovation Building (SCIB), 28 Kirkbride Lane, 
  Tuscaloosa, AL 35487-0288\\
  Email: mahmad12@crimson.ua.edu \\
  \hfill\break
  \textbf{Jiahe Cao}\\
  Department of School of Computing, University of Nebraska-Lincoln\\
  262 Avery Hall, 1144 T St, Lincoln 68588 Nebraska\\
  Email:jcao10@huskers.unl.edu\\
   \hfill\break%
  \textbf{Mizanur Rahman, Ph.D.}\\
  Department of Civil, Construction \& Environmental Engineering, The University of Alabama\\
  Smart Communities and Innovation Building (SCIB), 28 Kirkbride Lane, 
  Tuscaloosa, AL 35487-0288\\
  Email: mizan.rahman@ua.edu \\
  \hfill\break
  \textbf {Li Zhao, Ph.D}\\
  Department of Civil and Environmental Engineering, University of Nebraska-Lincoln\\
  262K Prem Paul Research Center at Whittier School\\
  2200 Vine Street, Lincoln 68588 Nebraska\\
  Email: lizhao@unl.edu, (402) 472-1928\\
  \hfill\break%
  \textbf {Nathan Huynh, Ph.D.}\\
  Department of Civil and Environmental Engineering, University of Nebraska-Lincoln\\
  262D Prem Paul Research Center at Whittier School\\
  2200 Vine Street, Lincoln 68588 Nebraska\\
  Email: nathan.huynh@unl.edu\\
  \hfill\break%
  \textbf {Eren Erman Ozguven, Ph.D.}\\
  Department of Civil and Environmental Engineering, FAMU-FSU College of Engineering\\
  2525 Pottsdamer Street, Tallahassee, FL, 32311\\
  Email: eozguven@eng.famu.fsu.edu\\
  \hfill\break%
}

\begin{document}
\maketitle

\section{Abstract}

Infrastructure-based sensing and real-time trajectory generation hold significant promise for improving safety in high-risk roadway segments like work zones, yet practical deployments are hindered by perspective distortion, complex geometry, occlusions, and costs. This study tackles these barriers by (i) integrating roadside camera and LiDAR sensors into a cosimulation environment to develop a scalable, cost-effective vehicle detection and localization framework, and (ii) employing a Kalman Filter-based late fusion strategy to enhance trajectory consistency and accuracy. In simulation, the fusion algorithm reduced longitudinal error by up to 70\% compared to individual sensors while preserving lateral accuracy within 1–3 meters. Field validation in an active work zone, using LiDAR, a radar-camera rig, and RTK-GPS as ground truth, demonstrated that the fused trajectories closely match real vehicle paths, even when single-sensor data are intermittent or degraded. These results confirm that KF based sensor fusion can reliably compensate for individual sensor limitations, providing precise and robust vehicle tracking capabilities. Our approach thus offers a practical pathway to deploy infrastructure-enabled multi-sensor systems for proactive safety measures in complex traffic environments.


\hfill\break%
\noindent\textit{Keywords}: work zone, fusion, lidar, camera, localization, safety
\newpage

\section{Introduction}
Work zone crashes do not necessarily impact only the vehicles and people directly involved; instead, they have cascading effects that cause operational delays for passing vehicles and project completion delays for work zone contractors. The Federal Motor Carrier Safety Administration (FMCSA) report indicates that commercial motor vehicles (CMVs) are involved in one-third of work zone fatal crashes, although they represent only 5\% of all vehicular traffic \cite{FMCSA2021}. Driving behavior (fatigued/aggressive/inattentive) is a major concern in these work zone crashes. In addition, speed is a contributing factor in 26\% of all fatal work zone crashes \cite{FDOT2022}. According to Jiao (2022) \cite{Jiao2022}, 13\% of CMV drivers are fatigued when they are involved in crashes. Driver fatigue is caused by extended driving hours, limited sleep, and/or drug use. Between 2018 and 2020, a consistent trend was observed in the number of intoxicated large truck drivers involved in fatal accidents ($\sim$100 events per year). The Associated General Contractors of America (AGC) conducts surveys to gather insights from private firms involved in managing work zones, focusing on their experiences and challenges. Based on survey responses collected between 2020 and 2022, phone usage and speeding were identified as the two main factors contributing to work zone crashes \cite{AGC2020, AGC2021, AGC2022}. Real-time vehicle trajectory data collection strategies to characterize the approaching vehicle stream (aggressive/fatigued) for work zones are adopted by a limited number of public agencies. Only 33\% of all state Departments of Transportation (DOTs) have implemented work zone data collection technologies \cite{BrownEdara2022}; 27\% have no plans to incorporate the use of data collection technologies in the future.

Infrastructure-based sensing and real-time vehicle trajectory generation are increasingly recognized as vital steps for enhancing safety in high-risk roadway segments, such as work zones. Traditionally, state Departments of Transportation (DOTs) have relied primarily on video cameras along with other traditional traffic sensors (loop detectors, pneumatic tubes, etc.) to collect traffic monitoring data. While video cameras can provide high-resolution images to detect and track objects in their field of view, their performance can be impacted by poor lighting and adverse weather conditions. Having redundant sensors, such as LiDAR, can help overcome these issues, as these sensors can operate effectively in various lighting and weather conditions. LiDAR provides the shape and orientation of objects when they are close to the sensor. Fusing data from multiple sensors can improve object detection and tracking accuracy, allowing for advanced warning of the approaching vehicle stream. This will further enhance work zone data quality compared to the existing standalone systems employed by agencies. Sensors, such as LiDAR and radar are increasingly being used in traffic control applications, as they provide comprehensive, real-time traffic data, including precise information on vehicle location, type, and trajectory \cite{li2023sensor}. Utilizing different sensors simultaneously to depict a scenario can provide complementary data, thereby overcoming the limitations of individual sensors and enabling more accurate information through sensor fusion.

While sensors like LiDAR, radar, and cameras offer promising capabilities for accurate vehicle detection and localization, their deployment in real-world settings remains limited. This is primarily due to technical challenges, including perspective distortion, complex road geometry, and vehicle occlusions, along with concerns related to cost and scalability in field deployments. To address these challenges, this study pursues two key objectives: (i) to develop and evaluate an infrastructure sensor-enabled vehicle detection and localization framework within a co-simulation environment for cost-effective and scalable evaluation; and (ii) to develop and evaluate a Kalman Filter (KF)-based late multi-sensor fusion (also known as decision level fusion) approach aimed at improving the consistency and reliability of vehicle trajectory generation. In this study, we conduct a simulation-based evaluation to demonstrate the effectiveness of real-time vehicle detection and localization using roadside camera and LiDAR sensors separately, as well as the KF-based late fusion of roadside LiDAR and camera data. Two separate simutators were used in this study: (1) SUMO (Simulation of Urban MObility) and (2) CARLA (Car Learning to Act). The microscopic traffic simulator, SUMO, helps to create realistic vehicle maneuvering behavior, whereas autonomous driving simulator, CARLA, helps to create realistic weather and driving conditions as well as roadside sensor data generation. To further validate the approach, we evaluated KF-based late fusion using a real-world work zone scenario with field-collected data from LiDAR, a radar-camera system, and a vehicle equipped with RTK-GPS.

\section{LITERATURE REVIEW}  
Roadside sensors offer broader spatial coverage with varying viewing angles and levels of reliability compared to onboard sensors \cite{lei2022integrated, chai2023roadside}. To fully leverage the advantages of individual sensors, it is essential to fuse information collected from multiple sources, enabling more accurate and comprehensive environmental perception \cite{chen2019multi}. Multi-modal fusion integrates data from heterogeneous sensors—such as cameras, LiDAR, and radar—and has been widely explored in the literature. Some fusion approaches utilize semantic information extracted from images and structural information from point clouds. Depending on the stage at which data are combined, sensor fusion can be categorized into three levels: sensor-level fusion (also known as early or data-level fusion), feature-level (mid-level) fusion, and decision-level (late) fusion.

Sensor-level fusion refers to the direct integration of raw data from multiple sensors. This approach benefits from rich datasets, potentially resulting in higher accuracy. However, handling raw data introduces challenges such as increased computational complexity and latency, which can hinder real-time decision-making. Despite these drawbacks, sensor-level fusion remains one of the most commonly applied techniques in transportation. For example, Xiang et al. \cite{xiang2022multi} propose a sensor-level perception method that fuses semantic data from cameras and range data from LiDAR at the data layer. Notable examples of sensor-level fusion methods include PointPainting \cite{vora2020pointpainting}, which projects LiDAR points onto the semantic segmentation output of camera images; MVX-Net \cite{sindagi2019mvx}; and PointAugmenting \cite{wang2021pointaugmenting}, both of which integrate multi-sensor data at the raw data level.

Feature-level fusion (mid-level fusion) occurs after each sensor processes its raw data into intermediate feature representations. These features are then fused to produce a unified perception output. This method reduces the computational burden associated with sensor-level fusion, though it presents challenges in aligning features from different sensor modalities—some information may be lost during the feature extraction stage. Anisha et al. \cite{anisha2023automated} fuse bounding box outputs from a low-resolution roadside camera and a solid-state LiDAR to extract data such as pixel localization, vehicle classification, and road user localization. Zhang et al. \cite{zhang2022robust} utilize an implicit feature pyramid network (i-FPN) to process multiscale image features under adverse conditions and introduce a hybrid attention mechanism (HAM) to mitigate distortions in point clouds and images caused by poor weather. Li et al. \cite{li2021enhancing} integrate 2D event pixels from an event-based camera—offering both spatial and temporal resolution—with 3D sparse point clouds from LiDAR at the feature level. Similarly, Zhao et al. \cite{zhao2023spatial} extract features from both camera imagery and LiDAR to estimate the geolocation of road users, achieving spatial alignment by fusing these modality-specific estimates across multiple views.

Decision-level fusion occurs when each sensor independently processes its input and outputs a detection result, which is then fused with outputs from other sensors to make a final decision. This approach is robust to individual sensor failures, as the system can still function using the results from operational sensors. Additionally, because it processes less data than sensor- or feature-level fusion, it is more computationally efficient. Wang et al. \cite{wang2022object} fuse 2D trajectory data from cameras with 3D trajectory data from LiDAR. Chai et al. \cite{chai2023roadside} use adaptive federated Kalman filtering to integrate vehicle trajectory data. Liu et al. \cite{liu2022object} employ an evidential architecture based on Dempster–Shafer theory \cite{dempster2008upper, shafer1976theory}, assigning different weights to radar and camera data. Da et al. \cite{da2019system} generate local maps using onboard sensors, which are then fused at a central node using the Independent Opinion Pool (IOP) algorithm \cite{pathak20073d}. Wang et al. \cite{wang2021roadside} propose a novel roadside fusion system combining camera and radar detections. Zheng et al. \cite{zheng2022robust} introduce RCP-MSF, a cooperative perception framework that fuses 2D and 3D bounding boxes from cameras and LiDAR using the Hungarian algorithm in combination with a Region of Interest (ROI) filter. Mo et al. \cite{mo2022method} fuse detections from infrastructure and vehicle-mounted sensors using an improved Kalman Filter.

In this study, we develop a decision-level (or late fusion) framework using a linear Kalman Filter to dynamically weight the outputs from roadside cameras and LiDAR sensors. This approach offers a computationally efficient alternative to data- or feature-level fusion by preserving robustness while avoiding the high processing overhead associated with raw data or feature matrix fusion.

\section{METHOD}
This section presents an infrastructure sensor-enabled vehicle detection and localization approach within a SUMO-CARLA co-simulation environment for cost-effective and scalable evaluation, as well as a KF-based late (or decision level) multi-sensor fusion approach to improving the consistency and reliability of vehicle trajectory generation.

\subsection{Object Detection and Localization in Co-Simulation Environment}

\subsubsection{Using Camera}
\begin{figure}[!htbp]
    \centering    \includegraphics[width=1\linewidth]{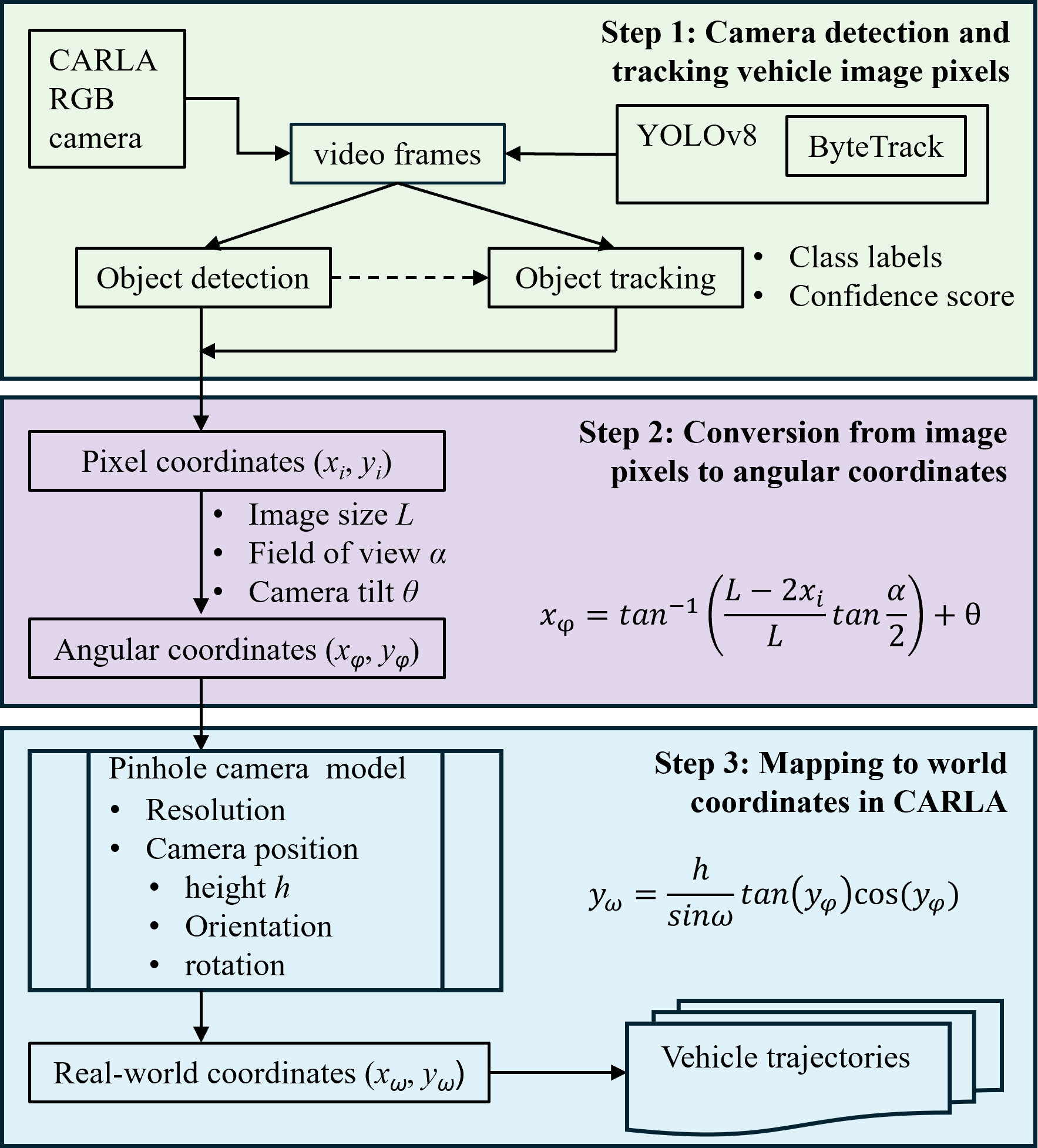}
    \caption{Camera-based object detection}
    \label{camera-detect}
\end{figure}

We use an RGB camera in CARLA for object detection. The camera is placed at a fixed location near the start of a work zone within the SUMO-CARLA co-simulation environment. The camera-based object detection pipeline captures real-time video frames at a certain rate (e.g., 10 frames per second). Each frame is processed using the YOLO (You Only Look Once) deep learning model, which detects dynamic objects (e.g., vehicles) and generates bounding boxes. To track the movement of the bounding boxes, detections are linked across frames where a unique ID is attached to each detected object. As the object moves, its trajectory is recorded in the image plane and then transformed into real-world coordinates using the pinhole camera model and the camera’s intrinsic parameters. The transformation relies on projecting 2D pixel coordinates into 3D world coordinates. Figure \ref{camera-detect} provides the flowchart of the camera-based object detection algorithm. The detection process includes three main steps, and they are described below. 
\noindent
\subsubsection{Step 1: Vehicle Detection and Tracking in CARLA}

\paragraph{\\
The RGB camera sensor in CARLA captures video frames at a defined resolution and frame rate. Each frame is processed using the YOLOv8 deep learning model to detect vehicles. YOLO generates bounding boxes, class labels (e.g., "car", "truck"), and confidence scores. Object tracking is then applied to associate detections across consecutive frames, which enables the construction of vehicle trajectories.\\} 

\subsubsection{Step 2: Conversion from Image to Angular Coordinates}
\paragraph{\\
The pixel coordinates P$(x_i, y_i)$ of each detected object are transformed into angular coordinates P$(x_\phi, y_\phi)$ through a relationship that accounts for the camera's field of view and image resolution. Specifically, each pixel position is mapped to an angle relative to the camera’s optical axis based on the horizontal and vertical angles subtended by the field of view. Using $x_i$ to $x_\phi$ as an example, the conversion can be done using Equation \ref{eq1}. In CARLA, the camera's intrinsic parameters are known, such as image resolution and size L, field of view $\alpha$ (default is 90 degrees horizontal), and camera tilt $\theta$. These allow precise calculation of the angular location of each object with respect to the camera's optical axis.}

\begin{equation}
    x_ \varphi = \tan^{-1} \left( \frac{L - 2x_i}{L} \tan \frac{\alpha}{2} \right) + \theta
    \label{eq1} 
\end{equation} 
\subsubsection{Step 3: Mapping to World Coordinates in CARLA}
\paragraph{\\
The camera's position, rotation, and orientation (height above ground and yaw) are also defined in the simulator. Using these camera configurations and angular relationships, we compute each object’s 3D world position using a pinhole camera model. Under the assumption that vehicles are located on a flat road \(z = 0\), the model can be expressed in Equation \ref{eq2}.}

\begin{equation}
y_\omega = \frac{h}{\sin \omega} tan(y_\varphi) cos(x_ \varphi)
\label{eq2} 
\end{equation}
\paragraph{
The transformation accounts for both camera rotation and placement in the co-simulation environment. This approach outputs trajectories in world coordinates, which are used to analyze vehicle behavior near the work zone. A validation process is then followed by comparing the estimated world coordinates (in simulation) with the ground truth trajectories obtained directly from CARLA and SUMO.}

\subsubsection{Using LiDAR}
\begin{figure}[!htbp]
    \centering
    \includegraphics[width=0.95\linewidth]{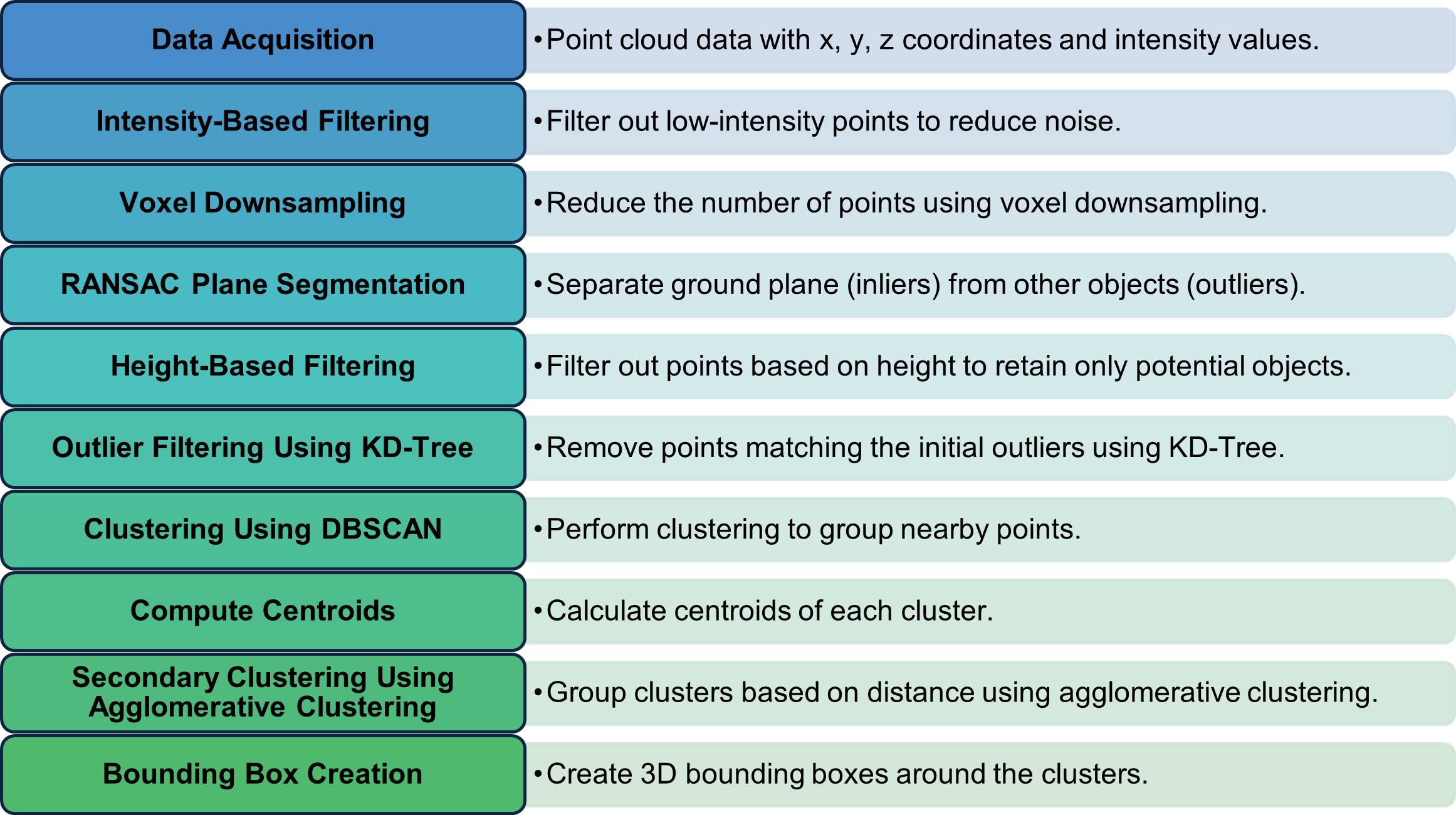}
    \caption{LiDAR-based Object Detection}
    \label{lidar-object-detection}
\end{figure}

In this study, several open-source frameworks for point cloud-based 3D object detection are reviewed, including Complex-YOLO \cite{simony2018complex}, TorchPoints3D \cite{simony2018complex}, and MMDetection3D\cite{mmdet3d2020}. MMDetection3D is studied for further experimentation due to its extensive model support and modular design. Inference is conducted with the PointPillars model which is trained on the KITTI 3D detection dataset. During simulation experimentation in this study, the model was unable to reliably detect approaching vehicles accurately. These prebuilt models are trained on the KITTI training dataset, which was collected from vehicle-mounted, ego-perspective sensors and lacks representation of fixed, elevated viewpoints. In the LiDAR‐based object detection pipeline as described in Figure \ref{lidar-object-detection}, a bottom-up paradigm of point cloud interpretation is proposed that involves background subtraction, followed by spatio-temporal clustering and classification \cite{leonard2008perception,himmelsbach2008lidar}. 

As described in Figure \ref{lidar-object-detection}, the very first step is LiDAR data acquisition and preprocessing the data. Data pre-processing begins by limiting the point cloud to a region of interest and removing obvious artifacts. An intensity-based filtering is applied to discard points with LiDAR return intensities below a threshold, thus reducing noise. A heuristic filtering like this is one of the most common first steps in LiDAR-based object detection \cite{duan2024small}. As raw point clouds contain millions of points, voxel downsampling is used to reduce the number of points while preserving the overall structure.

The road surface produces a large cluster of points that has the ability to merge with the object points. To separate ground points from obstacle points, a Random Sample Consensus (RANSAC) \cite{fischler1981random} plane-fitting method is used for ground segmentation in this study. After this step, a height-based filtering approach is applied to discard points outside a plausible height range, especially points that are above the sensor’s mounting height or below the road surface. The next step in the pipeline is removing isolated outliers using a k-d tree-based neighbor search.  For each point, the algorithm finds its k nearest neighbors and computes the average distance. Points that have unusually large average neighbor distances are labeled as noise and dropped. By discarding these outliers, the robustness of subsequent clustering can be improved, and thereby false positive detection can be reduced. 

After removing ground points, the scattered clusters of above-ground points are grouped together. Density-Based Spatial Clustering of Applications with Noise, DBSCAN \cite{ester1996density} is employed in this work. DBSCAN works by connecting points that are within a neighboring radius. In this method, points in high-density regions grow into clusters, while sparse noise points remain unassigned and are treated as outliers. The outcome of DBSCAN is a set of clusters, each representing a candidate object, and some noise points marked as noise.

For each point cluster, we compute the 3D coordinates of points in the cluster, often known as the geometric centroid.  This centroid serves as an estimated object center position in the scene. A secondary clustering approach is performed using Agglomerative Clustering in addition to the DBSCAN Clustering method. Agglomerative clustering operates in a bottom-up manner: each point starts as its own cluster, then pairs of clusters are successively merged based on proximity criteria. Euclidean distance between cluster centroids is used as the linkage metric, which can merge the two closest clusters at each step until reaching a stopping threshold. After clustering, we obtain a collection of point groups, each assumed to correspond to a single physical object. 

An oriented 3D bounding box is fitted around each cluster to enclose the object’s extent in the final step of the pipeline. Here, each 3D bounding box, along with center position, dimensions, and orientation, as well as its associated point cluster and centroid, represents a detected object.

\FloatBarrier

\subsubsection{LiDAR and Camera Data Processing}

An individual sensor can produce unreliable data for real-time localization. To reduce the error, a discrete Kalman Filter (KF) was implemented to smooth and estimate each vehicle's position over time. This processing is applied to the output of individual sensors, as raw localization output from sensors can provide noisy position measurements extracted from LiDAR point cloud clusters or camera bounding boxes. A constant velocity motion model is used, making the processing suitable for short-horizon driving scenarios where vehicle dynamics can be assumed to be locally linear.

The Kalman Filter assumes a constant-velocity motion model, with the state vector defined as:
\begin{equation}
\mathbf{x}_k = 
\begin{bmatrix}
x_k \\
y_k \\
\dot{x}_k \\
\dot{y}_k
\end{bmatrix}
\label{eq7}
\end{equation}

representing the 2D position ($x_k$, $y_k$) and velocity ($\dot{x}_k$, $\dot{y}_k$) of a detected object at timestep~$k$. The system gets updated according to the linear dynamical model:
\begin{equation}
\mathbf{x}_k = \mathbf{F}\mathbf{x}_{k-1} + \mathbf{w}_{k-1},
\quad \text{with} \quad
\mathbf{F} = 
\begin{bmatrix}
1 & 0 & \Delta t & 0 \\
0 & 1 & 0 & \Delta t \\
0 & 0 & 1 & 0 \\
0 & 0 & 0 & 1
\end{bmatrix}
\label{eq8}
\end{equation}

Here, \( \mathbf{w}_{k-1} \sim \mathcal{N}(0, \mathbf{Q}) \) is zero-mean Gaussian process noise, where the covariance matrix \( \mathbf{Q} \) is derived from the discretized white noise acceleration model:
\begin{equation}
\mathbf{Q} = \sigma_a^2
\begin{bmatrix}
\frac{\Delta t^4}{4} & 0 & \frac{\Delta t^3}{2} & 0 \\
0 & \frac{\Delta t^4}{4} & 0 & \frac{\Delta t^3}{2} \\
\frac{\Delta t^3}{2} & 0 & \Delta t^2 & 0 \\
0 & \frac{\Delta t^3}{2} & 0 & \Delta t^2
\end{bmatrix}
\label{eq9}
\end{equation}

Observations \( \mathbf{z}_k = [x_k, y_k]^\top \) from each sensor are related to the state as:
\begin{equation}
\mathbf{z}_k = \mathbf{H}\mathbf{x}_k + \mathbf{v}_k,
\quad \text{with} \quad
\mathbf{H} =
\begin{bmatrix}
1 & 0 & 0 & 0 \\
0 & 1 & 0 & 0
\end{bmatrix}
\label{eq10}
\end{equation}

where \( \mathbf{v}_k \sim \mathcal{N}(0, \mathbf{R}) \) is the measurement noise. Separate measurement noise covariances \( \mathbf{R} \) were defined for LiDAR and camera, reflecting their respective positional uncertainties.

This individual sensor specific data smoothing setup allowed the system to account for heterogeneous sensor characteristics and maintain sensor-specific tracking continuity.

\subsection{3.2 SENSOR FUSION}
In this study, a decision-level fusion approach is adopted. The detection outputs and trajectory data from both the LiDAR and camera sensors are combined using a KF-based method. For evaluation, data from both Camera and LiDAR was also fused together using a simple averaging method \cite{emami2021long}. Compared to the simple averaging, the KF key benefits such as optimal estimation and computational simplicity \cite{7001546}. Specifically, the linear Kalman filter serves as a recursive estimator that integrates a dynamic system model with noisy observations to yield an optimal estimate of the system’s state \cite{kalman1960new}.

At first, a nearest distance-based object matching algorithm is implemented to link the individual vehicle specific data from camera and LiDAR sensor. Later a linear Kalman filter \cite{kalman1960new} based fusion approach is leveraged  under a constant velocity model, which assumes that the motion of the object has a constant speed and heading remain unchanged between time steps. Data streams from multiple sensors are fused together by modeling the system’s state as a random vector that evolves according to a linear dynamical model and then incorporating measurements from each sensor as a noisy observation of that state into a single, more accurate estimate of a dynamic state, which is the velocity and position in this study.
We modeled the Kalman Filter based fusion initializing the system based on Equations \ref{eq7} and \ref{eq8}. The state transition matrix can propagete position by $ x \;\gets\; x + v_x\,\Delta t
$. A process noise Q is added that accounts for unmodeled accelerations.

After this initialization process, the Kalman Filter operates in two alternating steps—prediction and measurement update. In the prediction step, the Kalman Filter projects the object’s state from the previous time step forward to the current time, using the motion model. It also predicts the Uncertainty covariance P. Mathematically, the prediction step can be written as: 
\begin{equation}
\hat{x}_p = F \hat{x}
\label{eq:11}
\end{equation}

\begin{equation}
P_p = F P F^T + Q
\label{eq:12}
\end{equation}

Here, $\hat{x_p}$  is the predicted state, $\hat{x}$ is the current state, $P_p$ is the predicted uncertainty and $P$ is the current uncertainty.

In the measurement-update step, the Kalman Filter updates its predicted state based on the new measurements from the sensors, LiDAR, and Camera. At each frame, we receive LiDAR  and Camera feeds and stack them into a combined measurement vector z of  $[x,\,y,\,v_x,\,v_y]^\mathsf{T}]$. \\
The  difference between the actual measurement and the predicted measurement, the innovation is calculated:
\begin{equation}
y = z - H\hat{x}_p
\label{eq:13}
\end{equation}

Here, H is an observation model,  which maps the true state space into the observed space.\\
The covariance $S = H.P_p . H^T +R$ is also calculated, where R is the measurement noise covariance, representing the detection uncertainty. The Kalman filter then applies a Kalman gain $K$ to blend this innovation into the state estimate. The Kalman Gain can be written as: \\
\begin{equation}
K = P_p H^T S^{-1}
\label{eq:14}
\end{equation}
The correction state the Kalman Filter updates its estimate of the vehicle’s state using the Kalman gain and measurement residual. Mathematically it can be written as,\\
\begin{equation}
\hat{x} = \hat{x_p} + K y
\label{eq:15}
\end{equation}

\begin{equation}
P = (I - K H) P_p
\label{eq:16}
\end{equation}

We run two updates back to back: first, the camera measurement, and then the LiDAR measurement. Both sources were fused into one posterior. The Kalman gain naturally weights each sensor according to the uncertainty it possesses; essentially, the sensors with lower noise have more influence over the final estimate.

\section{Multi-sensor Data Generation}

\subsection {SUMO-CARLA Co-simulation Environment}
\begin{figure}[!htbp]
    \centering
    \begin{subfigure}[b]{0.8\linewidth}
        \centering
        \includegraphics[height=0.8in, width=\linewidth]{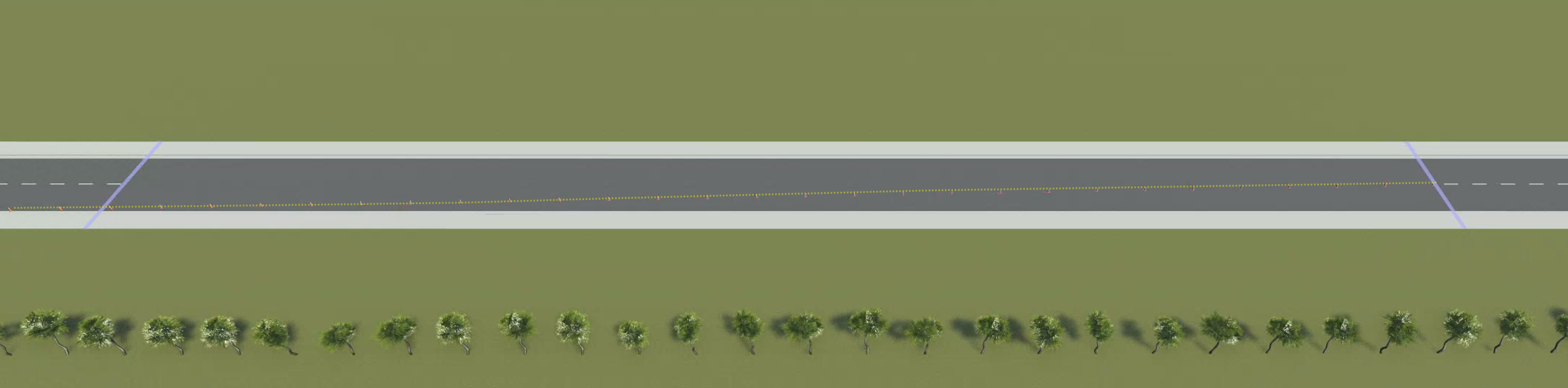}
        \caption{MUTCD compliant work zone setup in RoadRunner}
        \label{fig:workzone-setup}
    \end{subfigure}

    \vspace{1em} 

    \begin{subfigure}[b]{0.8\linewidth}
        \centering
        \includegraphics[height=2.2in, width=\linewidth]{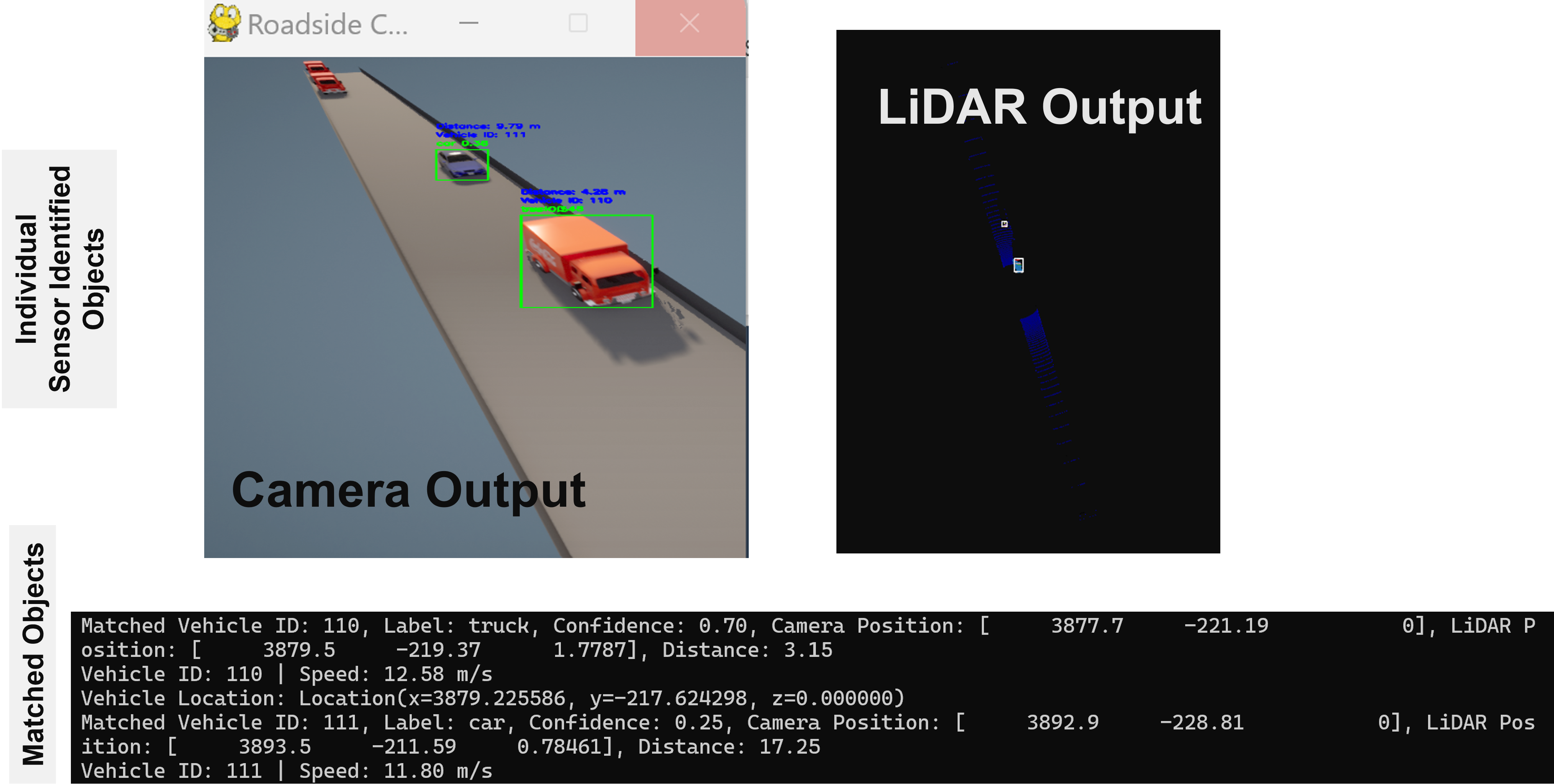}
        \caption{Simulated Camera and LiDAR based Vehicle Detection and Matching}
        \label{Camera- lidar-object-detection}
    \end{subfigure}

    \caption{Co-simulation environments and sensor outputs}
    \label{fig:combined-simulation}
\end{figure}

To simulate realistic sensor and vehicle behavior in a work zone scenario, a co-simulation framework is leveraged that integrates two simulators: CARLA and SUMO. SUMO is used to generate and route vehicle traffic using pre-configured trips, whereas CARLA provides a realistic sensor simulation. The test scenario involved a work zone where a two-lane road is reduced to a single lane, ensuring mandatory vehicle merging. At first, the Mathworks RoadRunner tool is used to setup the work zone scenario following the  Manual on Uniform Traffic Control Devices (MUTCD) guidelines, as shown in Figure \ref{fig:workzone-setup}. Later the network is imported in the xodr version to load in CARLA. Using CARLA's built-in functionality, the SUMO work zone network is generated. Vehicles are introduced at a fixed departure rate using SUMO route files and are randomly assigned to entry lanes. Sensor data is acquired using simulated sensors instantiated within CARLA. A 64-channel spinning LiDAR unit is deployed that has a 200-meter range, 3 million points per second density, and a 20 Hz rotation frequency. The field of view spans from 30 ° to + 11 °, ensuring comprehensive coverage of the merging zone. In addition, an RGB camera is mounted at the same location, configured with a 90-degree field of view and 1200×1200 resolution. These simulated sensors are placed upstream of the work zone area to replicate fixed roadside detection systems. All sensor data are collected in synchronization with the simulation timestep and processed in real time to extract vehicle object-level information. Figure \ref{lidar-object-detection} shows an examplescreenshot of the camera and LiDAR-based object detection and the nearest distance-based object matching output in real-time. 

\FloatBarrier

\subsection {Field Data Collection for Validation}
Field data were collected from an active work zone located at a bridge repair sites on N-2/US-75 in Lincoln, Nebraska, on June 23 and 24, 2025. The selected site featured a dynamic traffic condition where only one lane remained open to traffic at a time, resulting in a 2-to-1 lane drop configuration. This setup, along with a temporary reduced speed limit of 55 mph (down from the usual 70 mph), provided an ideal environment for studying real-world work zone behavior. To identify a suitable site, we coordinated with local agencies and contractors to evaluate potential locations along US-75. The chosen work zone offered the operational characteristics necessary to capture relevant traffic patterns and safety-related behaviors under constrained roadway conditions. Data acquisition was conducted using a high-resolution Ouster LiDAR sensor and an integrated Omnisight radar-camera system, both strategically deployed within the work zone. These sensing technologies enabled the collection of detailed, multimodal data on vehicle movement and interactions in the work zone, supporting subsequent analysis of traffic flow dynamics and safety assessments. For ground truth data collection, we equipped a minivan with a Single-antenna CPT7700 system, which is RTK GPS from NovAtel, a compact solution that combines a high-precision GNSS receiver with an inertial measurement unit (IMU). CPT7700 is also equipped with the high-performing Honeywell HG4930 Micro Electromechanical System (MEMS) IMU containing a gyroscope and an accelerometer. For the experiments, TerraStar-L correction services are utilized, and IMU data are logged at a 10 Hz frequency. 

\begin{figure}[!htbp]
    \centering    \includegraphics[width=0.95\linewidth]{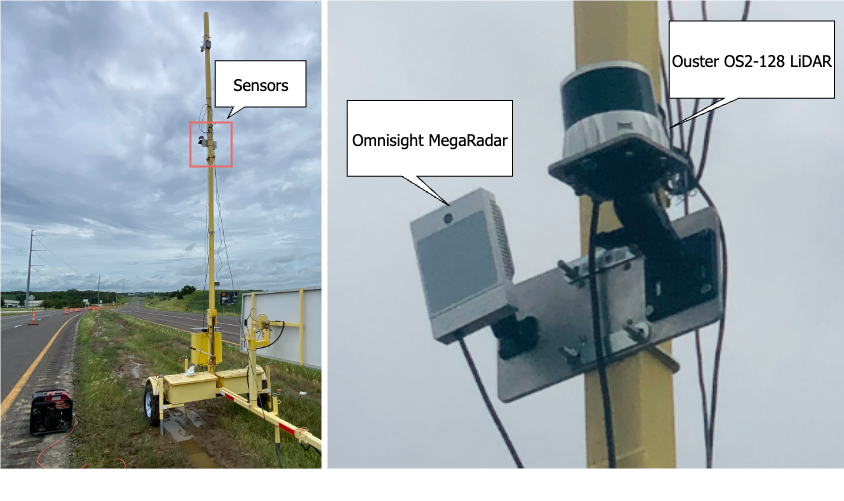}
    \caption{Ouster LiDAR sensor and an integrated Omnisight radar-camera system at work zone area on US-75}
    \label{field-lidar-radar-camera}
\end{figure}

\FloatBarrier

\section{RESULTS AND DISCUSSION}

In the following subsections, we have evaluated the effectiveness of a KF-based late multi-sensor fusion algorithm using data generated from the SUMO-CARLA co-simulation environment and field-collected data. 

\subsection{Evaluation of KF-based Late Fusion using Co-Simulation Data}
\begin{figure}[!htbp]
  \centering
  \hspace*{-0.1\linewidth}  
  \includegraphics[width=1.2\linewidth]{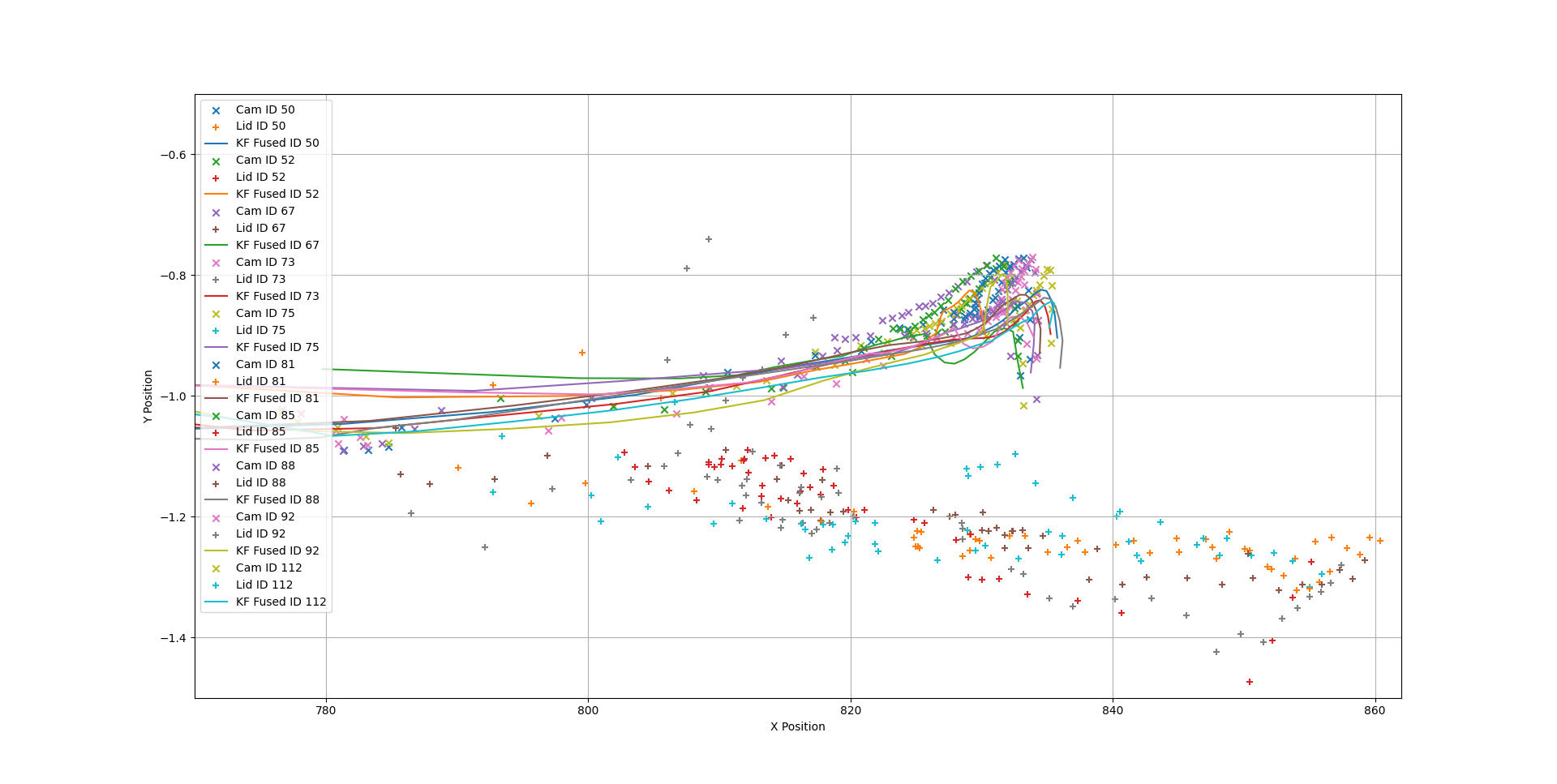}
  \caption{Kalman Milter Fused versus single sensor trajectories for multiple vehicles}
  \label{fig:kf_fusion}
\end{figure}

This subsection presents the findings from the multi-sensor fusion method for camera and LiDAR, focusing on the effectiveness of the localization method of the proposed algorithm. One hundred individual vehicles were simulated for this experimentation. Figure \ref{fig:kf_fusion} shows the longitudinal (x–axis) versus lateral (y–axis) positions for ten vehicles (IDs 50, 52, 67, 73, 75, 81, 85, 88, 92, and 112) as measured by the camera (markers), by the LiDAR (crosses), and as estimated by the Kalman‐filter fusion (solid lines). Noticeable jitter and occasional spikes are exhibited by both sensors. The Kalman Filter merges two data streams from Camera and LiDAR so that spurious deviations are strongly attenuated while preserving true lateral movements.

For simplicity in observing, we chose a small segment of the roadway between 818m to 833m. The localization error of both Kalman Filter and simple averaging methods were computed against the ground truth value for each of the vehicles, three specific scenarios were observed.\\

\subsubsection{Case 1: Fusion has better results than individual sensors}

\begin{figure}[!htbp]
  \centering
  \includegraphics[width=0.8\linewidth]{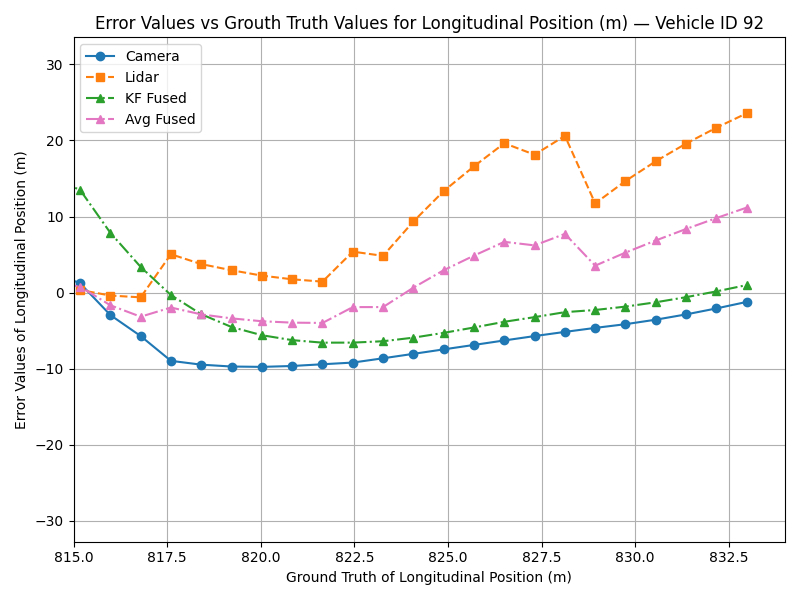} 
  \caption{Longitudinal‐error profiles for Vehicle 92}
  \label{fig:92}
\end{figure}

Figure \ref{fig:92} shows the error versus ground‑truth position for the camera (blue), LiDAR (orange), simple average (pink), and KF fusion (green). The camera error only starts with a large negative bias; it even reaches nearly 10m. After this point, it rises rapidly back toward zero as the vehicle moves from 815m to 833m. In contrast to that, the LiDAR exhibits a positive bias that grows from around $5m$ up to over $+20m$ by the end of the segment. Throughout the observed location, the KF fused estimated error values stay much closer to zero, balancing the opposing biases of the two sensors. Its peak error remains within $-5m $ across the entire observed trajectory. The simple averaging method, on the other hand, has an overall better error profile than individual sensors; however, it still lags behind the Kalman Filter-based fusion method in consistency. The fused solution outperforms both individual sensors, along with the simple averaging technique, by providing the smallest longitudinal error deviation for this scenario.

\subsubsection{Case 2: Camera has better results than the fusion method}

\begin{figure}[!htbp]
  \centering
  \includegraphics[width=0.8\linewidth]{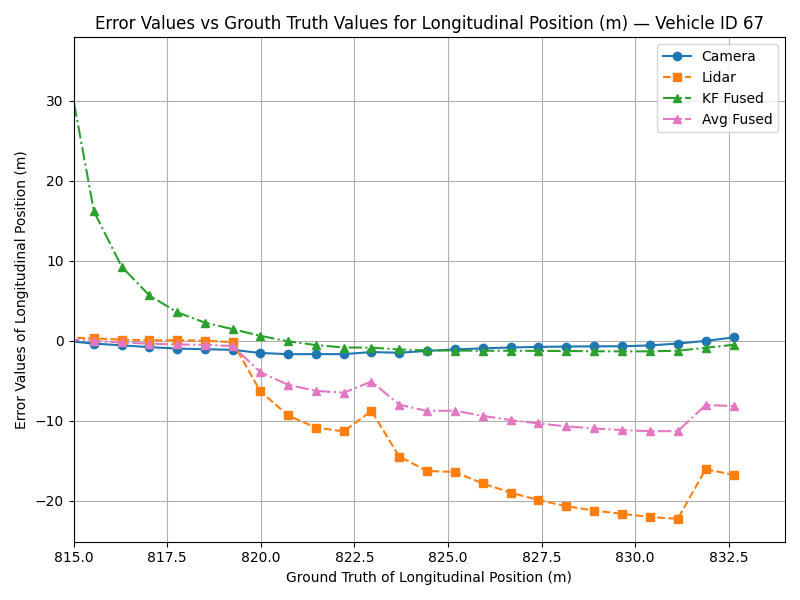} 
  \caption{Longitudinal‐error versus ground‐truth position for Vehicle 67}
  \label{fig:67}
\end{figure}

Figure \ref{fig:67} illustrates the camera-only errors (blue), LiDAR-only errors (orange), simple average (pink), and KF fusion (green). For this specific vehicle, the camera‐only error remains tightly clustered around zero, and only a slight negative drift between $818m$ and $820m$ can be observed, which did not even exceed $-1m$. Even though the LiDAR error starts near zero, it steadily worsens. The simple averaging mitigates some of the LiDAR bias; however, it still has errors in quite a high range in the later observed segment. On the other hand, the KF-based Fusion method corrects both sensor tendencies: it rapidly converges from an initial $+30m$ down to within $\pm1m$ of truth by $820m$ and stays centered around zero for the remainder of the trajectory. It is worth noting that the KF requires some time to initialize and calibrate, resulting in quite high initial error values. However, in general, the camera has a better error profile in this scenario than the fusion method.

\subsubsection{Case 3: Simple Averaging has better results than fusion method and individual sensors}

\begin{figure}[!htbp]
  \centering
  \includegraphics[width=0.8\linewidth]{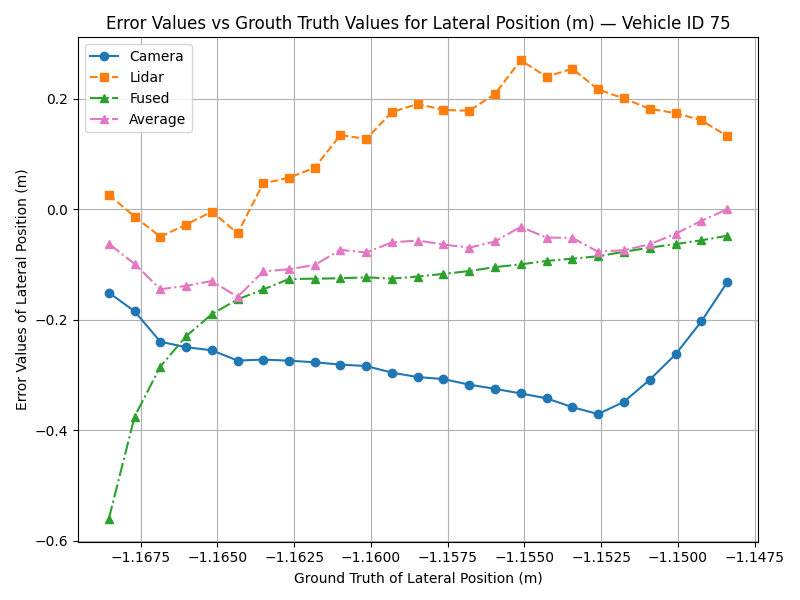} 
  \caption{Lateral‐error versus ground‐truth lateral position for Vehicle 88}
  \label{fig:75}
\end{figure}

\noindent
Figure \ref{fig:75} shows that the simple mean maintains the smallest overall deviation. The simple average stays closest to zero lateral error across the entire run. In contrast, the camera’s error steadily worsens. The LiDAR briefly has a good error profile; however, the error keeps increasing with the stretch. The KF Fusion result shows gradual improvement; however, it still lags behind the simple averaging values. Thus, in this case, the simple mean offers the most stable, least error‑prone lateral estimate. 

\begin{table}
\centering
\caption{Cumulative Absolute Errors for Longitudinal Distance}
\label{tab:cumulative-errors 1}
\begin{tabular}{ccccc}
\toprule
\textbf{Vehicle ID} & \textbf{Camera } & \textbf{LiDAR} & \textbf{KF Fused} & \textbf{Average} \\
\midrule
50  & 110.2078 & 329.2938 &  99.5990 & 219.7508 \\
52  &  80.3631 & 285.7167 &  40.8496 & 126.8989 \\
67  &  20.2712 & 291.7165 &  21.9895 & 155.5764 \\
73  & 113.5535 & 377.3983 & 101.4161  & 243.0096 \\
75  &  64.1556 & 304.0688 &  53.7558 & 182.0216 \\
81  &  90.9837 & 349.1209 &  83.2604  & 220.0523 \\
85  &  49.6368 & 266.3229 &  42.9169 & 157.9798 \\
88  & 113.5064 & 329.7424 &  98.9048 & 221.6244 \\
92  & 123.7249 & 228.5675 &  71.1760 &  95.8032 \\
112 & 125.5183 & 104.6122 &  87.0560 &  94.5296 \\
\bottomrule
\end{tabular}
\end{table}

\begin{table}[ht]
  \centering
  \caption{Cumulative Absolute Errors for Lateral Distance}
  \label{tab:table2}
  \begin{tabular}{ccccc}
    \toprule
    \textbf{Vehicle ID} & \textbf{Camera} & \textbf{LiDAR} & \textbf{KF Fused} & \textbf{Average} \\
    \midrule
     50 & 6.20006 &  1.81128 & 1.73223 & 2.19439 \\
     52 & 6.16245 &  0.65660 & 2.71595 & 3.23638 \\
     67 & 6.08377 & 10.09040 & 3.24318 & 3.91877 \\
     73 & 6.10020 &  1.46517 & 0.99617 & 2.52232 \\
     75 & 5.99901 &  3.11583 & 2.21754 & 1.48981 \\
     81 & 6.20139 &  1.99827 & 1.62250 & 2.10156 \\
     85 & 5.99782 &  6.65343 & 2.29112 & 2.44050 \\
     88 & 5.94356 &  2.06241 & 1.46627 & 1.94058 \\
     92 & 5.92250 &  2.34662 & 2.70800 & 3.86698 \\
    112 & 5.40375 &  1.35530 & 1.30082 & 2.05965 \\
    \bottomrule
  \end{tabular}
\end{table}

The total absolute longitudinal error for each vehicle was also calculated for the selected section of roadway for the individual sensors, KF-based late fusion, and equally weighted (i.e., average) late fusion method. Table \ref{tab:cumulative-errors 1} summarizes the total absolute longitudinal error accumulated between ground‑truth positions 818m and 833m for each of the ten selected vehicles. From Table \ref{tab:cumulative-errors 1}, it is observed that the KF fused method has a lower value of error than the average method and the LiDAR-only scenario for each vehicle. Only vehicle 67 has a lower value of cumulative absolute error for the camera than that of the fused method. Otherwise, the KF-based Fusion has achieved better results for the camera. Also, from Table \ref{tab:cumulative-errors 1}, Figures \ref{fig:67} and \ref{fig:92} show that the LiDAR exhibits very high error values in the longitudinal position. However, the fusion result is not affected by the poor performance of LiDAR. As the KF continuously estimates and adapts to each sensor’s uncertainty, it effectively reduces errors introduced by the LiDAR measurements whenever they exhibit large longitudinal errors, without discarding them entirely. 

Table \ref{tab:table2} represents the cumulative absolute error values for lateral distance for the selected vehicles and their trajectory path. It is observed that the camera only estimation performs worse in the lateral direction, the cumulative error values for each vehicle exhibit a relatively narrow error range of 5.4 m to 6.2, reflecting their consistent but biased performance. By contrast, the LiDAR-only errors vary more widely (0.66–10.09 m), driven by occasional depth‐measurement outliers. The KF fusion method outputs remain constrained within 0.99–3.24 m, demonstrating that the Kalman filter effectively mitigates extreme LiDAR noise while preserving the overall accuracy. The averaging method on the other hand fall between the individual sensor and fused values, highlighting the benefit of dynamic weighting in the KF fusion process over simple averaging. 

\FloatBarrier

\subsection{Evaluation using Field-collected Data}

\begin{figure}[!htbp]
  \centering
  \includegraphics[width=1\linewidth]{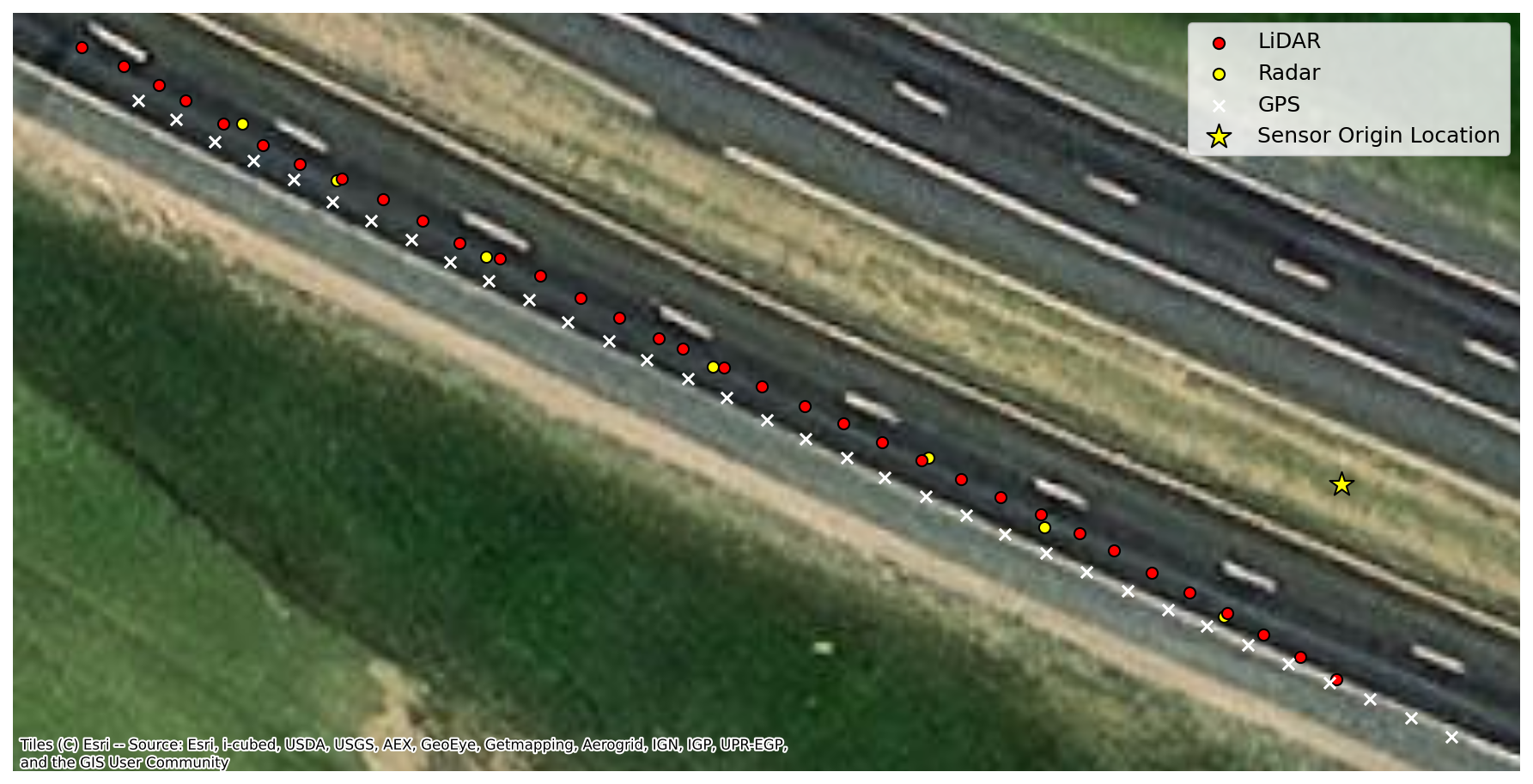} 
  \caption{Visualization of trajectories from LiDAR, camera-radar, and GPS (ground truth)}
  \label{fig:field-data}
\end{figure}

Figure \ref{fig:field-data} presents a comparative visualization of vehicle trajectories captured from LiDAR, camera-radar, and high-precision GPS (serving as ground truth) for a single test vehicle. Due to the limited availability of Real-Time Kinematic (RTK) GPS units, we were able to equip only one vehicle with this high-precision GPS system. For this vehicle, we simultaneously collected synchronized data from the LiDAR, camera-radar, and GPS sensors to evaluate the accuracy and performance of our sensor fusion approach. As observed in Figure \ref{fig:field-data}, the LiDAR sensor provided high-frequency vehicle trajectory data, while the camera-radar system produced trajectory points at a comparatively lower frequency. This deliberate use case was chosen to highlight the robustness and adaptability of our KF-based late fusion method. The primary objective was to assess how effectively our fusion approach can compensate for sensor limitations or data dropouts. It is necessary to demonstrate that when only one sensor provides measurements, the fusion framework seamlessly utilizes that single-source data in the measurement-update loop to maintain continuous vehicle tracking. Conversely, when both sensors are available, their data are jointly leveraged to enhance the accuracy of the estimated trajectory. This demonstration can prove that our KF-based late fusion method can reliably produce high-frequency, consistent trajectories even in scenarios with partial sensor availability or intermittent sensor failures, thereby improving system resilience in real-world deployments.

\begin{figure}[!htbp]
  \centering
  \includegraphics[width=1\linewidth]{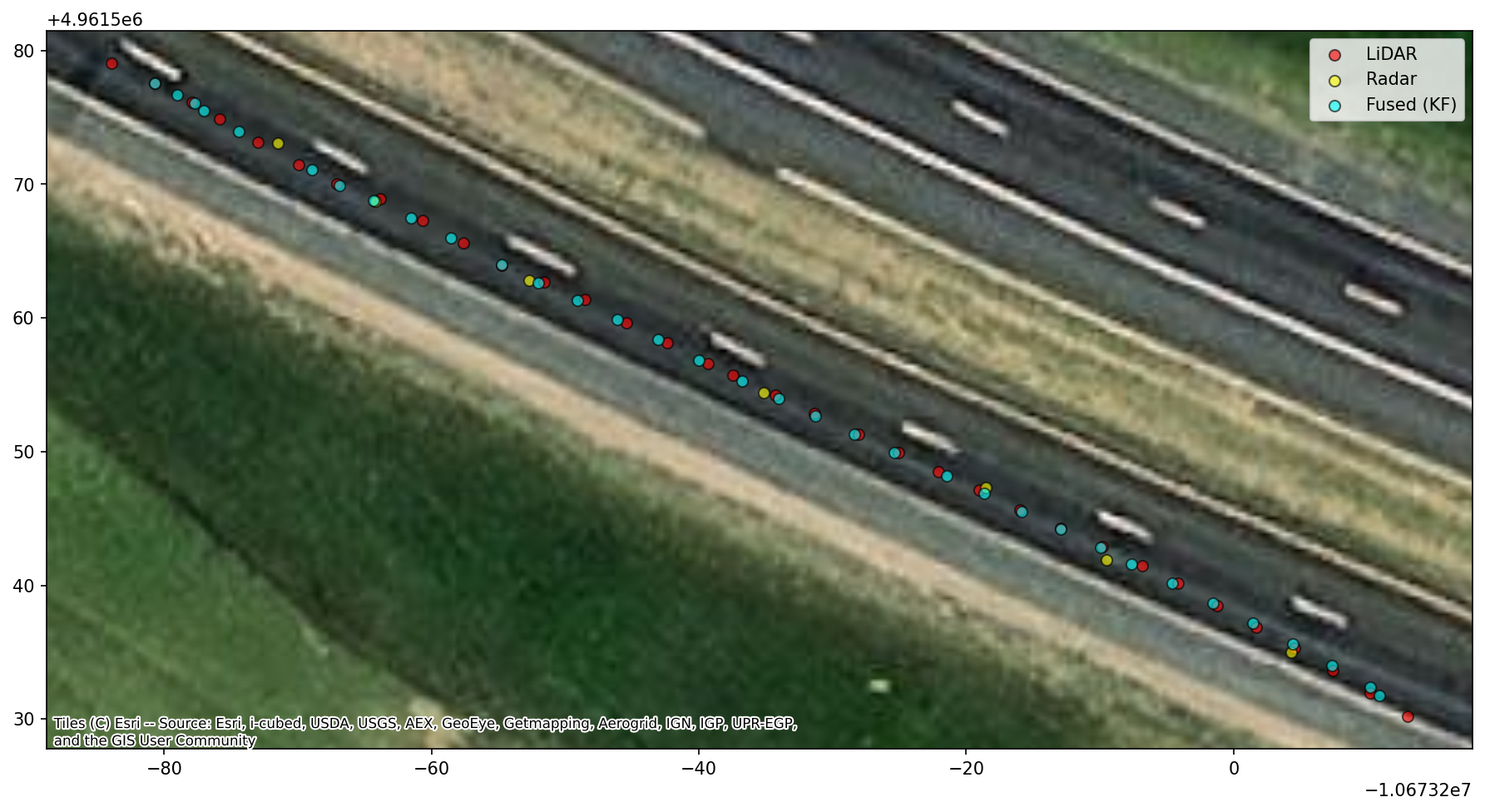} 
  \caption{Visualization of trajectories from LiDAR, camera-radar, and KF-based late fusion}
  \label{fig:kalman-traj-updated}
\end{figure}

To evaluate the effectiveness of our KF-based late fusion approach, we used field-collected data from LiDAR and a camera-radar sensor system. This method was designed to enhance the robustness and accuracy of vehicle trajectory estimation in a work zone area, as mentioned before. Figure \ref{fig:kalman-traj-updated} illustrates a comparative visualization of the individual trajectories obtained from the LiDAR and camera-radar sensors, alongside the trajectories generated through the KF-based late fusion process. The fused trajectories demonstrate close alignment with the ground truth data, indicating the effectiveness of the fusion technique in improving tracking accuracy. A key feature of our fusion algorithm is its adaptability to varying sensor availability. Specifically, when data is available from only one sensor—either the LiDAR or the camera-radar—the measurement-update step in the KF relies solely on that sensor's output. Conversely, when both sensor streams are available, data from both are incorporated into the fusion process. This design ensures consistency and continuity in vehicle tracking, even in instances where one of the sensors intermittently fails to detect the vehicle. It is also noteworthy that some of the fused trajectory points appear clustered in the initial frames. This behavior is attributed to the time required for the KF to converge and stabilize, particularly when starting from less-than-optimal initial conditions. Future improvements can be made by refining the initialization parameters of the filter to expedite convergence and further enhance the precision of early trajectory estimates.

\FloatBarrier

\section{CONCLUSION}
Infrastructure‐based sensing and real‐time trajectory generation hold significant promise for improving safety in high‐risk roadway segments like work zones, yet practical deployments are hindered by perspective distortion, complex geometry, occlusions, and costs. This study tackles these barriers by (i) integrating roadside camera and LiDAR sensors into a co‐simulation environment to develop a scalable, cost‐effective vehicle detection and localization framework, and (ii) employing a Kalman Filter–based late fusion strategy to enhance trajectory consistency and accuracy. In simulation, the fusion algorithm reduced longitudinal error by up to 70\% compared to individual sensors while preserving lateral accuracy within 1–3 meters. Field validation in an active work zone—using LiDAR, a radar–camera rig, and RTK‐GPS as ground truth—demonstrated that the fused trajectories closely match real vehicle paths, even when single‐sensor data are intermittent or degraded. These results confirm that KF‐based sensor fusion can reliably compensate for individual sensor limitations, providing precise and robust vehicle tracking capabilities. Our approach thus offers a practical pathway to deploy infrastructure‐enabled multi‐sensor systems for proactive safety measures in complex traffic environments.

\section{Funding}
This material is based upon work supported by the Federal Motor Carrier Safety Administration under grant FM-MHP-0792.
\section{Acknowledgments}
Any opinions, findings, and conclusions or recommendations expressed this publication are those of the author(s) and do not necessarily reflect the view of the Federal Motor Carrier Safety Administration and/or the U.S. Department of Transportation.

We used ChatGPT to help rephrase parts of our own writing to improve clarity and editorial quality.

\section{Athors Contribution}
\textbf{Suhala Rabab Saba, Sakib Khan and Minhaj Uddin Ahmad, Jiahe Cao:} conceptualization, coding, data collection, data analysis, methodology, and writing – original draft;  \textbf{Mizanur Rahman:} conceptualization, methodology, writing – original draft, and funding acquisition;  \textbf{Li Zhao, Nathan Huynh and Eren Erman Ozguven:} conceptualization, writing – review and editing, and funding acquisition.

\newpage

\bibliographystyle{trb}
\bibliography{trb_template}
\end{document}